\documentclass[11pt]{article}
\usepackage[a4paper, margin=1in]{geometry}
\usepackage{listings}
\usepackage{authblk}
\usepackage{fancyhdr}
\usepackage{hyperref}
\usepackage{amsmath}
\usepackage{amsfonts}
\usepackage{graphicx}
\usepackage{multirow}
\usepackage{multicol}
\usepackage{makecell}
\title{This part looks alike this: identifying important parts of explained instances and prototypes}

\author[1]{Jacek Karolczak\thanks{Corresponding author: \texttt{jacek.karolczak@cs.put.poznan.pl}}}
\author[1]{Jerzy Stefanowski}
\affil[1]{Poznan University of Technology, Institute of Computing Science, ul. Piotrowo 2, 60-695 Poznań, Poland}

\date{}

\begin{document}

\maketitle

\begin{abstract}
  Although prototype-based explanations provide a human-understandable way of representing model predictions they often fail to direct user attention to the most relevant features. We propose a novel approach to identify the most informative features within prototypes, termed alike parts.  Using feature importance scores derived from an agnostic explanation method, it emphasizes the most relevant overlapping features between an instance and its nearest prototype. Furthermore, the feature importance score is incorporated into the objective function of the prototype selection algorithms to promote global prototypes diversity. Through experiments on six benchmark datasets, we demonstrate that the proposed approach improves user comprehension while maintaining or even increasing predictive accuracy. 
\end{abstract}

\textbf{Keywords:} prototype-based explanation, feature importance, user attention guidance, local and global explanations

\section{Introduction}

Research on explaining black-box machine learning methods, which have been intensively developing in recent years,  has led to the introduction of a great number of various explanation methods; see, e.g.~\cite{bodria2023benchmarking}. Since prototypes correspond to training data, they are easier for humans to understand compared to more complex explanation methods~\cite{mastromichalakis2024prototypes}.  Prototypes can serve as a \textit{local explanation} by associating predictions with similar examples or as a \textit{global explanation} to illustrate model decision boundaries using a limited number of representative instances.

Although in general prototypes can be applied to different types of data, in this paper we focus on tabular data, i.e., the description of examples in the form of vectors of $\left(\text{feature}\,,\,\text{value}\right)$ pairs.  However, their interpretation may be a challenge, especially when there are too many features~\cite{mastromichalakis2024prototypes}. For local explanations in particular, human users may encounter difficulties in assessing which features are most important for the prediction of the considered instance. Furthermore, it can be expected for global explanations that the discovered prototypes are not only well spread over the learning data space but are simultaneously characterized by quite diversified subsets of the most important features.

Recall that similar expectations have been examined for other data modalities. For images, \textit{prototypical parts networks} were introduced to identify characteristic patches instead of complete images~\cite{chen2019protpnet}. However, for tabular data, the decomposition into meaningful parts remains underexplored. To bridge this gap, we introduce the identification of the most important features in prototypes. This is achieved by applying an agnostic explanation method for computing the feature importance of the black-box model, and offers a more refined perspective than existing techniques. Such subsets of features can be exploited for local or global approaches and support users in better interpreting the provided explanations. 

Our approach uses feature importance in two ways. First, we identify alike parts by highlighting the most informative overlapping features between an instance and its nearest prototype, directing the user's attention to a limited number of key features when interpreting a model prediction. Second, we incorporate feature importance into the prototype selection objective function to promote diversity, which aids in identifying alike parts. These strategies balance interpretability and diversity, enhancing both local explanations and prototype selection. The methods are evaluated on benchmark datasets, with source code available on GitHub\footnote{\url{https://github.com/jkarolczak/important-parts-of-prototypes}}.
\section{Related work}

A dataset $\mathcal{S}$ consists of $n$ instances (learning examples), expressed as $\mathcal{S} = {(\mathbf{x}_i, y_i)}_{i=1}^{n}$, where each $\mathbf{x}_i \in \mathbb{X}^d$ represents a $d$ dimensional feature vector, and $y_i \in \mathcal{Y}$ denotes its corresponding label. In this work, we consider tabular data in a feature-value format. We assume the presence of a classifier $h$ trained in $\mathcal{S}$, which serves as a black-box model to make predictions. The classifier $h$ maps an input instance $\mathbf{x}_i$ to a predicted label $y_i : h(\mathbf{x}_i) \mapsto \hat{y}_i$.  

From our perspective, a \textit{prototype} is a representative instance selected from the dataset, i.e., an element $(\mathbf{x}_j, \hat{y}_j)$, where $\hat{y}_j$ denotes the class assignment of the instance made by the classifier $h$. Typically, the set of prototypes, denoted as $\mathcal{P}$, is a subset of $\mathcal{S}$, such that $\mathcal{P} = \{(\mathbf{x}_j, y_j)\}_{j=1}^{m}$, where $m \ll n$ ($\mathcal{P} \subset \mathcal{S}$), ensuring that the number of prototypes is much smaller than the total size of the training dataset.

Some prototype selection methods use kernel functions and vector quantization~\cite{schleif2011kernel}, while KNN-based methods share similar principles. IKNN\_PSLFW~\cite{zhang2022distantprotos}, for example, partitions data into class-specific subsets and selects prototypes farthest from other classes. However, most methods rely on standard distance measures in the original attribute space, requiring a similarity definition that supports diverse data types (binary, numerical, categorical) and is robust to scaling differences. However, most of these algorithms exploit standard distance measures in the original attribute space, which requires the definition of similarity that supports different data types and is immune to different scales.

More recent proposals mitigate these distance limitations by considering the proximity of instances in the new space, referring to predictions of the black-box model; see the tree-space prototypes developed for explaining ensembles. The first algorithm, SM-A, introduced in \cite{tan2020tsp}, searches for prototypes -- medoids in this space. However, it requires the user to specify the expected number of prototypes. This limitation was later addressed by A-PETE \cite{karolczak2024apete}, which automates prototype selection.

Although numerous methods have been proposed to assess feature influence for black-box model predictions, they have not been widely applied in conjunction with prototype-based explanations. Popular techniques such as SHAP~\cite{lundberg2017shap} as a local explanation yield a vector of length equal to the number of features, where each value attributes the importance score of individual features, helping to understand the behavior of the model for specific instances.

Despite multiple studies on prototypes for tabular data, only a few papers discuss how prototypes should be presented to end users. In \cite{biehl2016prototypebasedmodels}, some prototype visualizations are provided, such as 2D scatter plots or self-organizing maps; however, they are suitable only for low-dimensional data and ultimately do not focus user attention on specific parts.

\section{Method}

In Section~\ref{subsec:important parts} we will first present our proposal to support the local explanation of the example predication by the nearest prototype. Then, in Section~\ref{subsec:new-optimization-problem} we will generalize it to create a diverse global set of prototypes.

\subsection{Identifying important parts}
\label{subsec:important parts}

\begin{table}[t]
    \centering
    \caption{Finding alike parts for the instance and its prototype from Apple Quality dataset. The first two rows present the feature importance values for the instance and its prototype, respectively. The third row shows the computed weights, obtained as the element-wise product of normalized feature importance scores (Formula 2). The bottom row indicates the binary mask, which selects the most relevant shared features-those with weights above the mean - denoted by '1'}
    \label{tab:mask-example}
    \begin{tabular}{lccccccc}
        \hline
         & Size & Weight & Sweetness & Crunchiness & Juiciness & Ripeness & Acidity \\
        \hline
        Instance & -2.77 & -1.08 & -1.72 & 1.38 & 0.19 & 3.65 & 0.31 \\
        Prototype & -0.97 & -0.20 & -3.07 & 0.00 & -0.52 & 3.16 & -0.52 \\
        Weights & 0.18 & 0.02 & 0.27 & 0.00 & 0.00 & 0.51 & 0.00 \\
        Mask & 1 & 0 & 1 & 0 & 0 & 1 & 0 \\
        \hline
    \end{tabular}
\end{table}

Following \cite{mastromichalakis2024prototypes}, for many features, a prototype as a whole can be difficult to comprehend and therefore make it difficult to explain the prediction of a black-box model. Some features within the prototype may be of high importance, while others may have low importance to the specific prediction that is being explained. 

Therefore, we propose a method that identifies the most informative features shared between an instance and its prototype, guiding the user's attention to a concise subset of features. Further, we refer to them as \textit{alike parts}, where the importance of features within the alike part is similarly high in both the instance and its nearest prototype.

To explain the instance $\mathbf{x}_i$ by its nearest prototype $\mathbf{p}_j$, we first identify the alike parts by computing feature importance scores for each feature $l \in {1, \dots, d}$ in the classification of $\mathbf{x}_i$ and $\mathbf{p}_j$ using the classifier $h$, denoted as $\phi(h, \mathbf{x}_i^l)$ and $\phi(h, \mathbf{p}_j^l)$, respectively. We use the SHAP method \cite{lundberg2017shap} to quantify the influence of each feature, as it is one of the most widely used methods for feature importance estimation. However, any feature importance method can be applied in this context. To ensure comparability, the raw importance scores are normalized, as they can vary in magnitude. We treat both positive and negative scores equally by squaring them, which avoids cancellations and enables the identification of similarities and differences between the instance and its prototype:

\begin{equation}
    \hat{\phi}(h, \mathbf{x}_i^l) = \frac{(\phi(h, \mathbf{x}_i^l))^2}{\sum_{k=1}^{d} (\phi(h, \mathbf{x}_i^k))^2}, \quad
    \hat{\phi}(h, \mathbf{p}_j^l) = \frac{(\phi(h, \mathbf{p}_j^l))^2}{\sum_{k=1}^{d} (\phi(h, \mathbf{p}_j^k))^2}\,.
\end{equation}

To quantify the alignment of feature importance between the instance and the prototype, we define a weight for each feature as the product of its normalized importance scores:

\begin{equation}
    w_l = \hat{\phi}(h, \mathbf{x}_i^l) \cdot \hat{\phi}(h, \mathbf{p}_j^l)\,.
\end{equation}

These weights are used to determine the degree to which each feature highly influences the prediction of the model for both the prototype and the explained instance.  Various operators can achieve this - here we propose to select a subset of the most influential features -- by defining a binary mask $\mathbf{m} \in \{0,1\}^d$, where these features with weights above the mean of all values are retained:

\begin{equation}
    m_l = \mathbb{1}\left( w_l > \frac{1}{d} \sum_{k=1}^{d} w_k \right)
\end{equation}

Table~\ref{tab:mask-example} illustrates the identification of a subset of important features.

\subsection{New definition of optimization problem}
\label{subsec:new-optimization-problem}

The prototype selection algorithms discussed in this paper, such as \cite{tan2020tsp,karolczak2024apete}, define the task of identifying representative data points as a $k$-medoids problem, which is solved using a greedy approximation algorithm. Typically, the $k$-medoids problem minimizes a distance function $d$ between each training example $\mathbf{x}_i$ and its nearest prototype~$\mathbf{p}_j$. This is expressed as follows:

\begin{equation} 
    f(\mathcal{P}) = \sum_{i=1}^{|\mathcal{S}|} \min_{\mathbf{p}_j \in \mathcal{P}} d \left( \mathbf{x}_i, \mathbf{p}_j \right),
\end{equation}

\noindent where the notation $|\mathcal{S}|$ refers to the cardinality of the training set. The choice of the distance function $d$ varies between different algorithms. In neural network-based approaches, it can be a dot product between trainable embeddings \cite{li2018protonetwork}, or in tree ensembles, a specialized tree distance metric \cite{tan2020tsp,karolczak2024apete}.

To strengthen diversification in feature importance, we propose extending the objective function by including an additional feature importance component $fi$ defined as the product of normalized feature importance of $l$-th feature of instance~$\mathbf{x}_i$ and its nearest prototype~$\mathbf{p}_j$:

\begin{equation}
    fi(\mathbf{x}_i, \mathbf{p}_j) = \sum_{l=1}^{d} \frac{(\phi(h, \mathbf{x}_i^l))^2}{\sum_{k=1}^{d} (\phi(h, \mathbf{x}_i^k))^2} \cdot \frac{(\phi(h, \mathbf{p}_j^l))^2}{\sum_{k=1}^{d} (\phi(h, \mathbf{p}_j^k))^2}.
\end{equation}

The $fi$ scores can be calculated once for all $\mathbf{x}_i$ prior to optimization and cached for efficiency. The revised function incorporates both the minimization of the distance between each instance and its nearest prototype and an additional term weighted by $\beta$ to account for the feature importance score. The first term promotes that each instance in the dataset is well represented by a prototype, promoting compact coverage of $\mathcal{S}$ by assigning each instance to its closest prototype, while the second encourages diversification in the feature importance across prototypes. The revised function is formally defined as:

\begin{equation} 
    f(\mathcal{P}) = \sum_{i=1}^{|\mathcal{S}|} \min_{\mathbf{p} \in \mathcal{P}_j} \left( d \left( \mathbf{x}_i, \mathbf{p}_j \right) + \beta \cdot fi \left( \mathbf{x}_i, \mathbf{p}_j \right) \right),
\end{equation}

This modification enables a more nuanced global prototype selection, with $\beta$ balancing distance and feature importance. The updated formulation improves prototype selection for identifying alike parts.

The proposed method is robust to missing values, assuming that the selected components can handle them. In this paper, we used prototype selection algorithms \cite{tan2020tsp,karolczak2024apete} based on RF \cite{breiman2001rf}, and SHAP, both of which natively support missing values. Therefore, the method does not require additional preprocessing for missing data.
\section{Experiments}

\begin{table}[t]
    \caption{An example of an instance and its alike parts identified from the nearest prototype using the A-Pete algorithm \cite{karolczak2024apete}. The selection is based on two optimization problem definitions: the original (raw) and the Feature Importance (FI)-informed approach. Parts alike between the explained instance and prototype in the FI-informed approach are bolded, while those alike in the original (raw) strategy are underlined.}
    \label{example-diabetes}
    \centering
    \begin{tabular}{lccccccccc}
    \hline
    type & Pregnancies & Glucose & BloodP. & SkinT. & Insulin & BMI & PedigreeF. & Age \\
    \hline
    instance & 6 & \underline{\textbf{102}} & 82 & 0 & 0 & 30.8 & \textbf{0.18} & \textbf{36} \\
    prototype (FI) & 7 & \textbf{125} & 86 & 0 & 0 & 37.6 & \textbf{0.30} & \textbf{51} \\
    prototype (Raw) & 7 & \underline{62} & 78 & 0 & 0 & 32.6 & 0.39 & 41 \\
    \hline
    instance & \underline{\textbf{8}} & \underline{\textbf{100}} & 74 & 40 & 215 & 39.4 & \textbf{0.66} & 43 \\
    prototype (FI) & \textbf{9} & \textbf{152} & 78 & 34 & 171 & 34.2 & \textbf{0.89} & 33 \\
    prototype (Raw) & \underline{9} & \underline{171} & 110 & 24 & 240 & 45.4 & 0.72 & 54 \\
    \hline
    \end{tabular}
\end{table}

As discussed in Section~\ref{subsec:new-optimization-problem}, the proposed optimization method can be adapted to various algorithms. We applied this modification to prototype selection algorithms optimizing tree distance: A-PETE, SM-A, and G-KM \cite{tan2020tsp,karolczak2024apete}, to explain the Random Forest (RF) ensemble~\cite{breiman2001rf}. All use greedy medoid selection, with key differences: G-KM selects an equal number of prototypes per class (greedy k-Medoid approximation computed within classes); SM-A \cite{tan2020tsp} selects the prototype providing the greatest improvement across all classes; and A-Pete \cite{karolczak2024apete} automates this by stopping based on relative improvements (see~\cite{karolczak2024apete} for pseudo codes). For evaluation, we use four benchmark datasets that have a subset of globally important features: Australia Rain\footnote{\url{https://www.kaggle.com/datasets/jsphyg/weather-dataset-rattle-package}}, Breast Cancer\footnote{\url{https://www.kaggle.com/datasets/rahmasleam/breast-cancer}}, Diabetes\footnote{\url{https://www.kaggle.com/datasets/mathchi/diabetes-data-set}}, and Passenger Satisfaction\footnote{\url{https://www.kaggle.com/datasets/teejmahal20/airline-passenger-satisfaction}}; and two: Apple Quality\footnote{\url{https://www.kaggle.com/datasets/nelgiriyewithana/apple-quality}} and Wine Quality\footnote{\url{https://www.kaggle.com/datasets/taweilo/wine-quality-dataset-balanced-classification}}, which exhibits high feature importance across all features.

The experiments are organized as follows: Section~\ref{subsec:results-intro} presents examples of alike parts identification on real data and how extending the optimized function improves this process. Section~\ref{subsection:comparison} aims to quantify the quality of the proposed improvements by comparing our modified with the original prototype selection methods, highlighting the impact of our changes. Section~\ref{subsec:ablation-study} presents an ablation study that analyzes the contribution of the $\beta$ factor to algorithm performance.

\subsection{Studying the methods in action}
\label{subsec:results-intro}

Finding alike parts on real data is shown in Table \ref{tab:mask-example}, illustrating how feature importance for both the instance and prototype is used to compute weights. Table \ref{example-diabetes} compares how alike parts of an instance and its nearest prototype are selected using the original (raw) and FI-informed versions of the A-Pete for the Diabetes dataset. Incorporating feature importance into A-Pete’s optimization led to different selections than the raw algorithm when generating prototypes from black-box RF \cite{breiman2001rf}.

For example, when using the prototype from raw A-PETE, only the \textit{Glucose} is highlighted as the feature important for both the instance and prototype. Meanwhile, the FI-informed algorithm also highlights \textit{Diabetes Pedigree Function}, and \textit{Age} which aligns with established medical knowledge on diabetes risk factors \cite{kautzky2016diabetes}. This demonstrates the potential of our method to facilitate the identification of more meaningful relationships between instances and prototypes.

\begin{figure}[t]
    \centering
    \includegraphics[width=\textwidth]{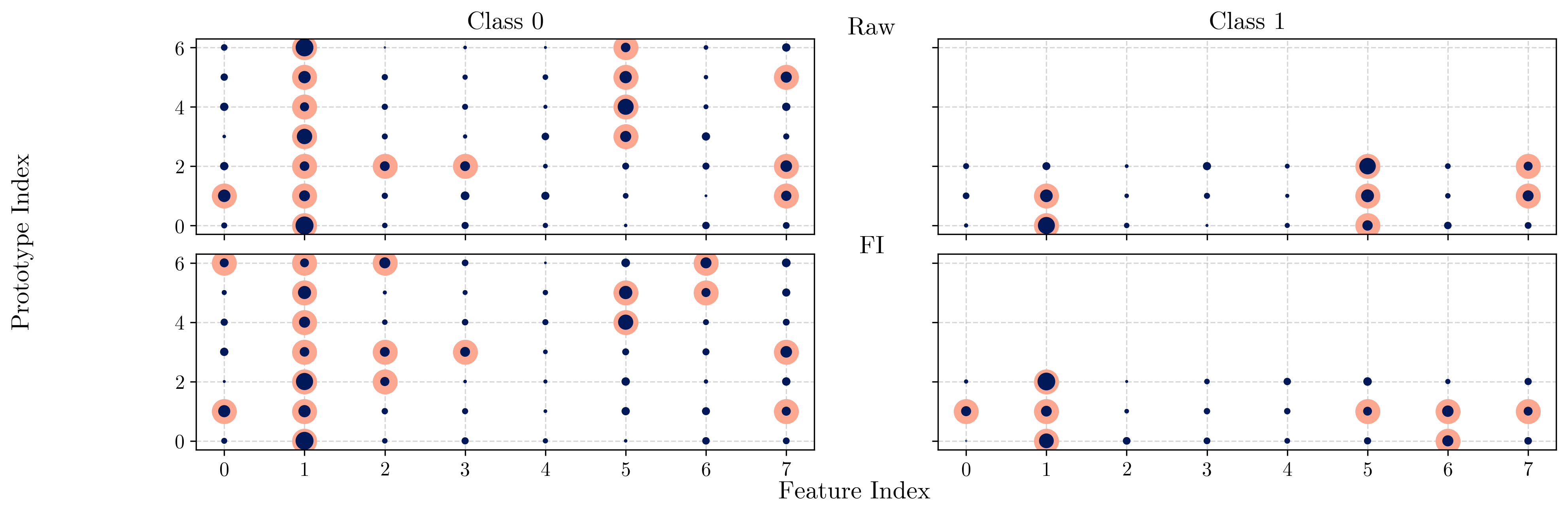}
    \caption{Comparison of prototypes (x-axis -- prototype index) and important features (y-axis -- feature index) for the Diabetes dataset. The top row displays prototypes generated using the original raw algorithm, while the bottom row incorporates an extended target function with feature importance (FI). The size of the inner circle represents feature importance, and pink highlights features identified as important for a given prototype.} 
    \label{fig:prototypes-comparison}
\end{figure}

A visual comparison of the globally generated sets of prototypes and selected important attributes for the Diabetes dataset is presented as Figure~\ref{fig:prototypes-comparison}. The figure contrasts prototypes generated using the original (raw) A-Pete algorithm with those generated using the FI-informed approach. The figure demonstrates that the FI-informed algorithm yields more diversified prototypes that highlight parts varying between prototypes -- the sixth feature was selected as important only when FI was included in the target function. A similar phenomenon was observed for Australia Rain and Breast Cancer -- certain features were considered significant only when using the FI-informed version of the algorithm.

\begin{figure}[t]
    \centering
    \includegraphics[width=\textwidth]{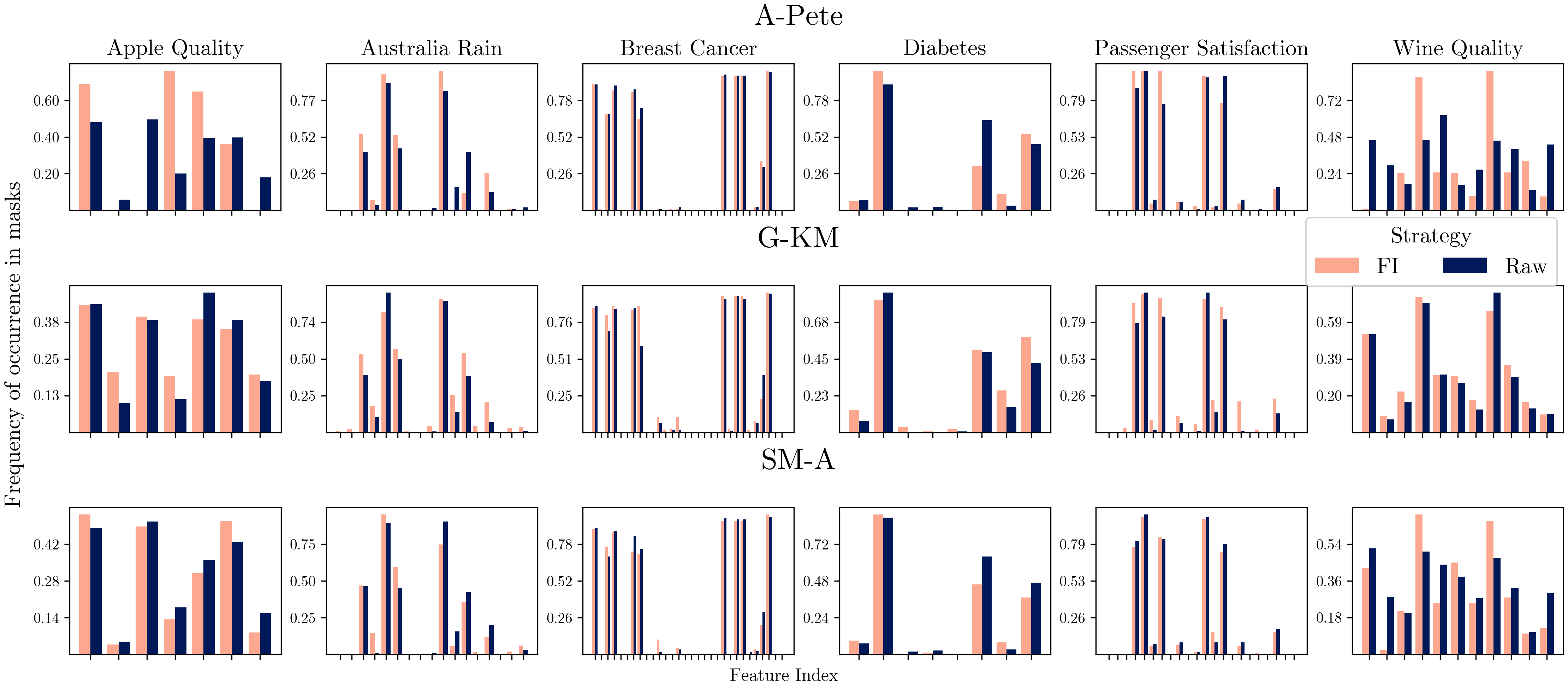}
    \caption{The comparison of the frequency of feature highlighting between the original (raw) and Feature Importance (FI)-informed strategies across different benchmark datasets. The results are shown for three prototype selection algorithms: A-Pete, G-KM, and SM-A.} 
    \label{fig:activating-features}
\end{figure}

The proposed approach was validated on the test subset of each dataset to quantitatively compare the frequency of features identified as important. In the Figure 2, presenting results, one can observe that the frequency of highlighting each feature differs between the original and FI-informed strategies. This difference is especially noticeable for the G-KM algorithm: when prototypes are selected using the FI-informed strategy, certain features are highlighted that were not emphasized by the raw algorithm.

\subsection{Predictive performance in comparison to original versions of the algorithms}
\label{subsection:comparison}

This section compares the accuracy achieved by a surrogate model based on prototypes, as it was done in \cite{tan2020tsp,karolczak2024apete}. The surrogate model uses a 1-nearest neighbor (1-NN) search within the set of selected prototypes and is evaluated on classifying instances from a test set.  We specifically examine the impact of our modified prototype selection method, which incorporates feature importance.

\begin{figure}[t]
    \centering
    \includegraphics[width=\textwidth]{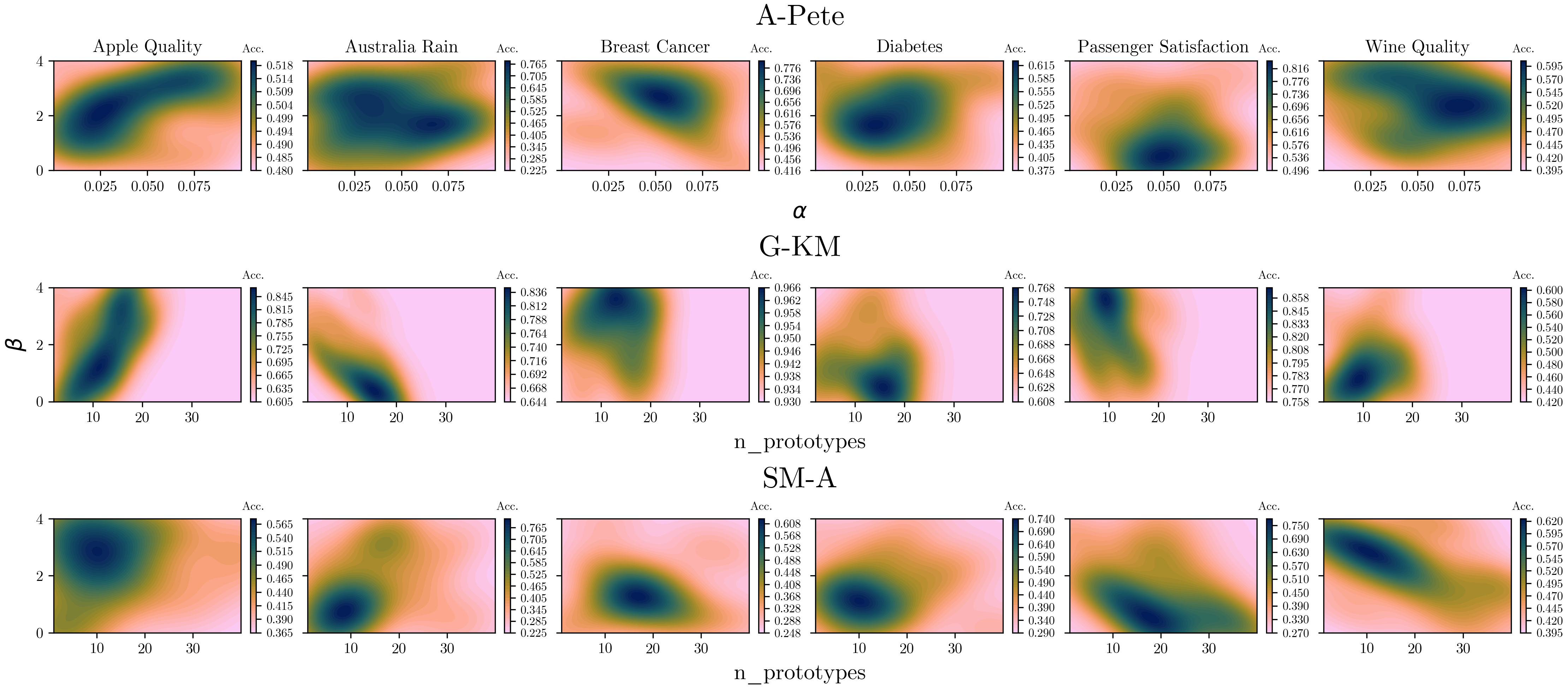}
    \caption{The comparison of accuracy (hue) achieved by A-Pete, G-KM, and SM-A algorithms across benchmarks against algorithm-specific hyperparameters (x-axis) and $\beta$ (y-axis). Note that the bottom line of each subfigure ($\beta = 0$) represents the original definition of the algorithms, where only the tree distance is minimized.} 
    \label{fig:accuracy}
\end{figure}

Figure~\ref{fig:accuracy} illustrates how algorithm-specific hyperparameters and the weighting factor $\beta$ influence prototype selection and consequently impact accuracy, with $\beta$ controlling the extent to which feature importance is incorporated into the optimization function. The results show that the modified approach maintains or improves predictive performance with respect to main parameters. Similar information is presented in Table~\ref{tab:accuracy} where the values corresponding to the accuracy optima found are presented for the original and the FI-incorporated algorithms.

\begin{table}[t]
    \centering
    \caption{Comparison of accuracy achieved by A-Pete, G-KM, and SM-A across benchmarks. The hyperparameters selected for the Feature Importance-informed version of the algorithm correspond to the maxima of accuracy in Figure~\ref{fig:accuracy}.}
    \label{tab:accuracy}
    \begin{tabular}{lc|cccccc}
        \hline
        \multirow{3}{*}{Algorithm} & \multirow{3}{*}{\makecell[l]{Objective\\function}} & \multicolumn{6}{c}{Dataset} \\
        & & \makecell[l]{Apple\\Quality} & \makecell{Australia\\Rain} & \makecell{Breast\\Cancer} & Diabetes & \makecell{Passenger \\ Satisfaction} & \makecell{Wine\\Quality} \\
        \hline
        \multirow[c]{2}{*}{A-Pete} & FI & 0.520 & 0.767 & 0.798 & 0.623 & 0.837 & 0.605 \\
        & Raw & 0.487 & 0.424 & 0.488 & 0.427 & 0.783 & 0.438 \\
        \hline
        \multirow[c]{2}{*}{G-KM} & FI & 0.861 & 0.843 & 0.965 & 0.766 & 0.865 & 0.602 \\
        & Raw & 0.785 & 0.822 & 0.939 & 0.739 & 0.781 & 0.541 \\
        \hline
        \multirow[c]{2}{*}{SM-A} & FI & 0.571 & 0.809 & 0.623 & 0.734 & 0.779 & 0.624 \\
        & Raw & 0.461 & 0.625 & 0.344 & 0.492 & 0.712 & 0.448 \\
        \hline
    \end{tabular}
\end{table}

\subsection{Ablation study}
\label{subsec:ablation-study}

\begin{figure}[t]
    \centering
    \includegraphics[width=\textwidth]{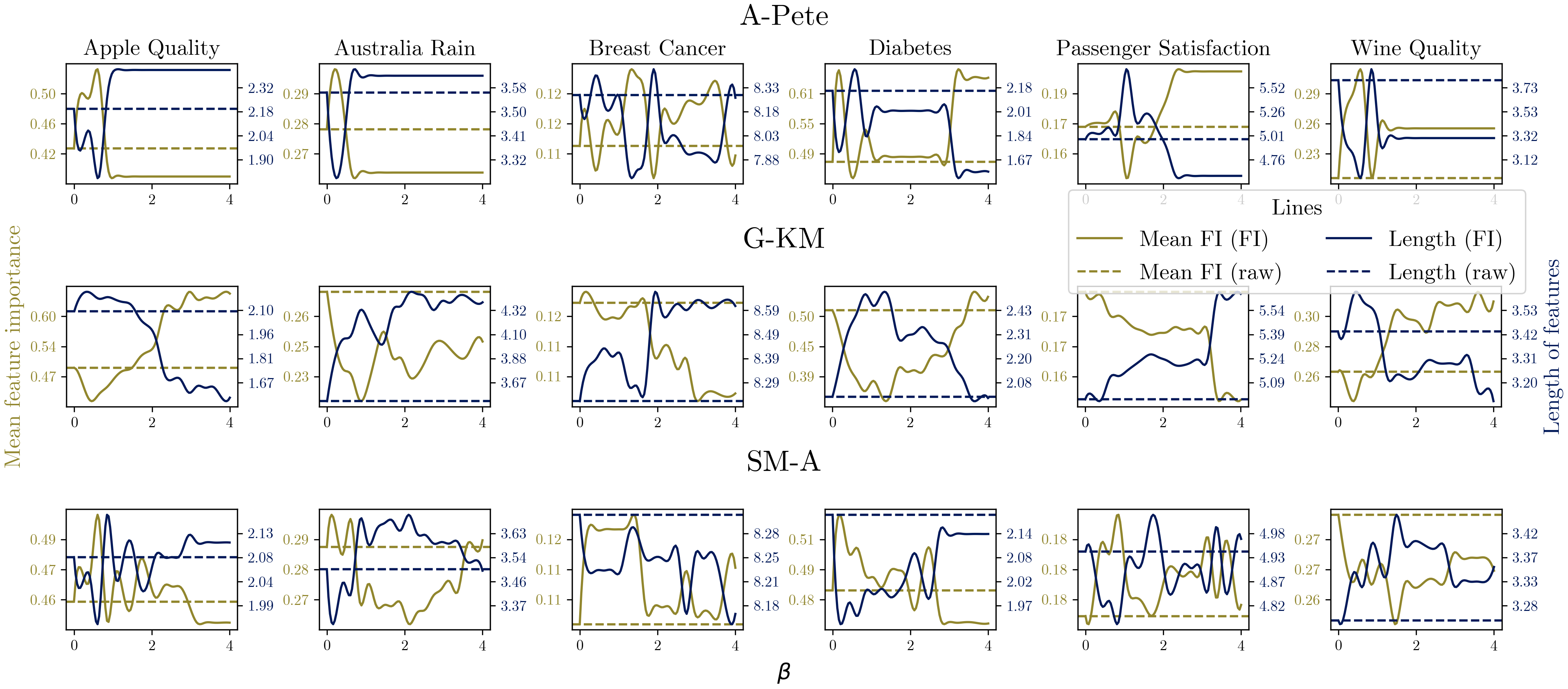}
    \caption{The comparison of mean feature importance of the features included in alike parts (left y-axis) and the length of the vector identified as alike parts between the explained instance and the prototype (right y-axis). The plot illustrates these two values tested against different $\beta$ values (x-axis).} 
    \label{fig:splines}
\end{figure}

Here, we analyze the impact of the parameter $\beta$ on the selection of the prototype by examining how it influences the alikeness between an explained instance and its prototype. Figure~\ref{fig:splines} shows how mean feature importance and alike-part length vary with $\beta$.

The results indicate that as $\beta$ increases, the mean feature importance similarity tends to rise, suggesting that high $\beta$ encourages the selection of prototypes that align more closely with important features of the explained instance. However, this trend is not strictly monotonic and careful tuning is required, with $\beta \leq 2.0$ often providing a good balance, although the optimal value depends on the dataset.

To determine the optimal value of $\beta$, grid search or Bayesian optimization can be used to tune $\beta$ and other algorithm-specific parameters, aiming to maximize the accuracy of a surrogate 1-NN model.
\section{Discussion}

This work introduces an innovative approach to prototype-based explanations, enhancing their interpretability by directing user attention to the most important features of both the prototype and the classified instance, the so-called alike parts. By incorporating feature importance into the prototype selection, our proposal bridges a gap in the literature where these two aspects were previously considered separately. The experimental results suggest that this integration improves the clarity of the explanation while preserving and, in some cases, even improving the predictive accuracy (see Section~\ref{subsection:comparison}).

Incorporating feature importance leads to selecting prototypes with different, often more meaningful, alike parts. This was shown with the Diabetes dataset, where our method identified features such as \textit{Age} and \textit{Pedigree Function} as crucial, aligning with established medical knowledge (see Section~\ref{subsec:results-intro}).

Moreover, it can extend beyond the tested algorithms, G-KM, SM-A, A-PETE, and a black-box RF. Importantly, Section~\ref{subsec:ablation-study} shows that adjusting the weighting factor $\beta$ fine-tunes the balance between feature importance and distance minimization, highlighting adaptability to different tasks.

Future research should explore its effectiveness from the user perspective, assessing whether these explanations enhance human understanding of model decisions. Furthermore, evaluating the approach on non-tabular modalities, such as images and text, is necessary to assess its broader applicability.

\section*{Acknowledgments}
This research was funded in part by National Science Centre, Poland OPUS grant no. 2023/51/B/ST6/00545 and in part by PUT SBAD 0311/SBAD/0752 grant.

\bibliographystyle{plain}
\bibliography{bibliography}

\end{document}